 \documentclass[pmlr,twocolumn,10pt]{jmlr} 

\usepackage{enumitem}
\usepackage{booktabs}
\usepackage{multirow}
\usepackage{siunitx}

\usepackage[utf8]{inputenc} 
\usepackage[T1]{fontenc}    
\usepackage{hyperref}       
\usepackage{url}            
\usepackage{booktabs}       
\usepackage{amsfonts}       
\usepackage{nicefrac}       
\usepackage{microtype}      
\usepackage{cleveref}       
\usepackage{lipsum}         
\usepackage{graphicx}
\usepackage{natbib}
\usepackage{doi}

\usepackage[frozencache,cachedir=.]{minted}

\newcommand{\equal}[1]{{\hypersetup{linkcolor=black}\thanks{#1}}}

\newcommand\reader{\texttt{meds\_reader}}

\newcommand\code{\url{https://github.com/som-shahlab/meds_reader_paper_code}}

\jmlrvolume{LEAVE UNSET}
\jmlryear{2024}
\jmlrsubmitted{LEAVE UNSET}
\jmlrpublished{LEAVE UNSET}
\jmlrworkshop{Machine Learning for Health (ML4H) 2024} %

\title[\reader{}]{\reader{}: A fast and efficient EHR processing library}

\author{%
\Name{Ethan Steinberg}\equal{These authors contributed equally} \Email{ethan.steinberg@gmail.com}\\
\addr Prealize Health and Stanford University, United States
\AND
\Name{Michael Wornow}\footnotemark[1] \Email{mwornow@stanford.edu}\\
\addr Stanford University, United States
\AND
\Name{Suhana Bedi}\footnotemark[1] \Email{suhana@stanford.edu}\\
\addr  Stanford University, United States
\AND
\Name{Jason Alan Fries} \Email{jfries@stanford.edu}\\
\addr  Stanford University, United States
\AND
\Name{Matthew B. A. McDermott} \Email{matthew\_mcdermott@hms.harvard.edu}\\
\addr  Harvard Medical School, United States
\AND
\Name{Nigam Shah} \Email{nigam@stanford.edu}\\
\addr  Stanford Healthcare and Stanford University, United States
}

\begin{document}

\maketitle

\begin{abstract}
The growing demand for machine learning in healthcare requires processing increasingly large electronic health record (EHR) datasets, but existing pipelines are not computationally efficient or scalable. In this paper, we introduce \reader{}, an optimized Python package for efficient EHR data processing that is designed to take advantage of many intrinsic properties of EHR data for improved speed. We then demonstrate the benefits of \reader{} by reimplementing key components of two major EHR processing pipelines, achieving 10-100x improvements in memory, speed, and disk usage. The code for \reader{} can be found at \url{https://github.com/som-shahlab/meds_reader}.

\end{abstract}

\paragraph*{Data and Code Availability}
We use three publicly available electronic health record datasets: eICU, MIMIC-III and MIMIC-IV \citep{eicu, mimiciii, mimiciv}. Our paper code can be found at \code{}.

\paragraph*{Institutional Review Board (IRB)}
We use de-identified data and thus did not require IRB approval.

\section{Introduction}
\label{sec:intro}


\begin{figure*}[!hbtp]
    \caption{An illustration of the event-stream nature of EHR data, showing an example subject and a visit within that subject's timeline.}
    \label{f:example}
    \includegraphics[width=\textwidth]{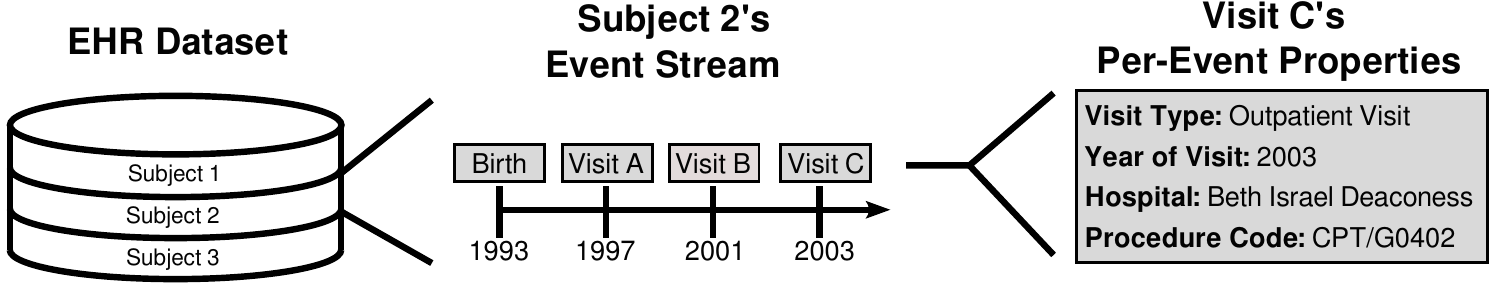}
\end{figure*}

As machine learning (ML) matures in both healthcare and other fields, there is a need to process large datasets for model training, especially with the rise of data-hungry foundation models \citep{scaling}. To meet growing data needs, sophisticated and efficient tooling \citep{sentencepiece, torchvision, datasets} have been developed to help scale analysis to large datasets. However, many of these tools have been difficult to use in research involving electronic health record (EHR) data due to its unique nested event stream data structure \citep{esgpt} consisting of a collection of subjects (also generally referred to as patients), where each subject contains a sequence of discrete time-stamped events with associated per-event data. Figure \ref{f:example} illustrates this event stream structure for an example subject. This event stream data structure is poorly handled by existing data processing tools that are optimized for tabular data, images, or text. These differences have forced healthcare ML researchers to build their own data processing pipelines \citep{pyhealth, pipeline, FIDDLE, esgpt} for handling EHR data, which tend to be very inefficient in terms of memory, CPU, and disk usage. 

In this work, we help with these inefficiency issues by introducing \reader{}, an open-source Python package that can be used for building fast and efficient EHR ML processing pipelines. We demonstrate the benefits of \reader{} by using it to reimplement labeling and featurization within two existing EHR processing pipelines, achieving 10-100x improvements in CPU, memory, and disk usage.

\section{Related Work}
\label{sec:related_work}

EHR ML researchers have developed many open-source EHR processing pipelines \citep{esgpt, pyhealth, FIDDLE, pipeline}. All rely on one of two main strategies for processing EHR data: pure Python or Python with many operations implemented using tabular libraries such as pandas. The goal of \reader{} is to enable the creation of more efficient EHR processing pipelines by offering an alternative strategy where data operations can instead be implemented using the event stream API provided by \reader{}.

\textbf{Pure Python:} Packages such as PyHealth \citep{pyhealth} convert the entire subject data into Python objects, using pickle to load and store data as necessary. Regular Python logic (such as for loops and if statements) can then be used to perform arbitrary analysis on these converted objects. This approach is fast (as Python conversions are not performed while processing) but requires high amounts of memory and disk due to Python object overhead.

\begin{figure*}[!htbp]
\caption{Example \reader{} transformation to find all subjects who have experienced stroke}
    \label{code.1}
    \begin{minted}{Python}
def find_stroke(subjects):
    subject_ids = set()
    for subject in subjects:
        for event in subject.events:
            if event.code.startswith('ICD10/I63'):
                subject_ids.add(subject.subject_id)
    return subject_ids

subject_database = meds_reader.SubjectDatabase('...', num_threads=10)
all_subjects = set()
for partial_result in subject_database.map(find_stroke):
    all_subjects |= partial_result
    \end{minted}
\end{figure*}

\textbf{Python with Tabular Libraries:} Packages such as EventStreamGPT \citep{esgpt}, FIDDLE \citep{FIDDLE} and MIMIC-IV Data Pipeline \citep{pipeline} use an alternative strategy where they offload processing tasks from Python to a variety of generic tabular data processing tools such as PySpark, pandas, and polars. These tabular data processing tools have drastically different implementations and design philosophies, but operate under the same principle of providing a tabular data abstraction where data consists of independent rows with specified columns. Python processing pipelines can then be written to operate on the table as a whole, and the tool will automatically use optimized Java, C++, or Rust to process every row. The main issue with this approach is that only some operations are natively supported by these libraries, often forcing EHR processing pipelines to perform complex transformations by converting data to Python, which comes at a significant cost.

\section{Design \& Implementation}

\reader{} is designed around the idea that electronic health record data is event stream data that consists of a collection of subjects, where each subject contains a sequence of time-stamped event data. EHR processing pipelines can then be written as a series of transformation steps of those per-subject event streams. Figure \ref{code.1} shows the code for one example transformation that identifies all patients who have experienced a stroke. \reader{} uses a variety of optimizations to make these transformations as fast and memory-efficient as possible.

\textbf{Event Stream Optimizations}: The key optimizations in \reader{} leverage the event stream nature of EHR data. First, \reader{} uses the subject as the primary unit for data storage and analysis, where all events for a subject are stored and processed together. Since many EHR data transformations require processing all events for a subject at once, \reader{} loads and caches all events for a subject whenever any event is accessed. This approach enhances processing speed by enabling the use of vectorization and allowing us to store events consecutively on disk and in memory, which results in better utilization of disk and memory caches. Second, \reader{} takes advantage of the time-oriented structure of the data by sorting events chronologically and processing events in order from the earliest to the latest time-stamp. Sorting by time enables the use of $O(n)$ incremental algorithms for featurization and labeling where only the new information at each timestep needs to be processed, as opposed to other algorithms that have to constantly reanalyze the whole record.

\textbf{EHR Specific Optimizations}: \reader{} also takes advantage of several common properties of EHR data to speed up operations. First, \reader{} exploits the repetitive nature of EHR data by caching common text strings. EHR data contains thousands of repeated copies of the same diagnosis codes and labs. \reader{} avoids Python object creation costs by only creating a single Python string object for every unique text value. Second, \reader{} optimizes for the sparse nature of EHR data using a mix of columnar storage and bitmaps. EHR data is sparse in that even though there are many possible properties for a given event (such as \texttt{units}, \texttt{discharge\_reason}, etc.), only a handful are ever filled out or processed in any given transformation. In order to handle transformations that only touch a few properties at a time, \reader{} uses columnar storage so that pipelines can only load the properties they need and ignore unnecessary ones. Per-event sparseness is handled using bitmaps, so unused property entries only take 1 bit each and locating the used entries can be done using fast $O(1)$ bitmap operations.

\textbf{Supporting a Variety of Datasets}: EHR datasets come in various formats and structures. \reader{} acknowledges that diversity, and supports arbitrary EHR data by taking advantage of the flexible Medical Event Data Standard (MEDS) \citep{meds}. MEDS provides a generic schema that defines EHR data as an event stream with arbitrary per-event properties. Using \reader{} requires converting a particular EHR dataset to MEDS, either using existing ETL packages, such as the publicly available ones for MIMIC-IV, eICU, or OMOP \citep{medsetl, medstransform}, or writing your own conversion logic. The simple structure of MEDS makes it straightforward to write custom code for these ETLs, which we do as part of our benchmarks on PyHealth and MIMIC-IV Data Pipeline (see Appendix \ref{a:convert}).

\label{sec:overview} 
\begin{figure*}[!htbp]
    \caption{An overview of how to use \reader{}, starting with converting to MEDS and ending with using a \reader{} SubjectDatabase to implement labeling and featurization.}
    \label{f:overview}
    \includegraphics[width=\textwidth]{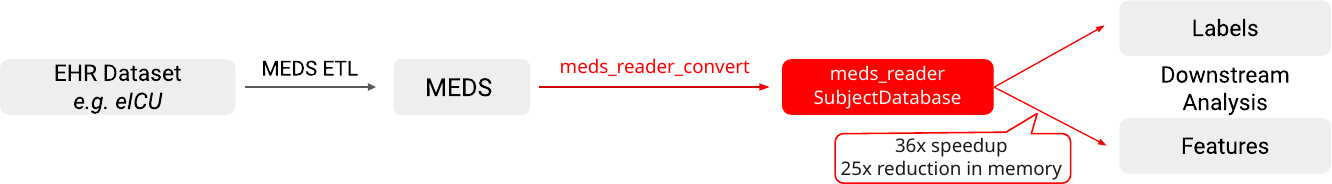}
\end{figure*}

\textbf{Python API And Usage}: \reader{} is implemented as three main Python classes: \texttt{Event}, which represents individual measurements in EHR data; \texttt{Subject}, which contains all events for a single subject; and \texttt{SubjectDatabase}, which manages collections of subjects for efficient querying and processing. A complete overview of the API for this package can be found at \url{https://meds-reader.readthedocs.io/en/latest/api_reference.html}. In order to use this API, it is necessary to first convert your dataset to MEDS, and then to the \reader \texttt{SubjectDatabase}, before finally using the resulting database for labeling and featurization (see Figure \ref{sec:overview}).

\section{Benchmarks}
\label{sec:benchmarks}

\begin{table*}[!htbp]
\small
\setlength{\tabcolsep}{4.5pt}
\begin{tabular}{ccc*{3}{r}*{3}{r}*{3}{r}}
\toprule
\multirow{2}{*}{Dataset} & \multirow{2}{*}{\#Subjects} & \multirow{2}{*}{Library} & \multicolumn{3}{c}{Memory (GB) $\downarrow$} &  \multicolumn{3}{c}{CPU (secs) $\downarrow$} & \multicolumn{3}{c}{Disk (GB) $\downarrow$} \\

\cmidrule(lr){4-6} \cmidrule(lr){7-9} \cmidrule(lr){10-12}

& & & Orig. & Ours & $\Delta\times$ & Orig. & Ours & $\Delta\times$ & Orig. & Ours & $\Delta\times$ \\

\midrule

eICU & 200,859  & PyHealth & 16.80 & \textbf{0.67} & 25.1$\times$ & 70.44 & \textbf{1.91} & 36.9$\times$ & 1.4 & \textbf{0.095} & 14.7$\times$ \\
MIMIC-III & 46,520  & PyHealth & 20.97 & \textbf{0.54} & 38.8$\times$ & 57.55 & \textbf{1.58} & 36.4$\times$ & 1.8 & \textbf{0.042} & 42.9$\times$ \\
MIMIC-IV & 299,712  & MDP & 14.32 & \textbf{2.09} & 6.9$\times$ & 293,261 & \textbf{7.44} & 39.4k$\times$ & 0.64 & \textbf{0.33} & 1.9$\times$ \\

\bottomrule

\end{tabular}
\caption{Compute resources used by various data processing pipelines compared to our reimplementations using \reader{}. We measure peak memory usage, wall clock CPU time, and the disk space used for cache directories. We compare the original resources used to the reimplementation and report the multiplicative change ($\Delta\times$). Orig. = Original pipeline. MDP = MIMIC-IV Data Pipeline. }
\label{t:results}
\end{table*}

The goal of \reader{} is to enable the creation of efficient EHR processing pipelines. To test whether it can achieve that purpose, we have reimplemented two components of two existing EHR processing pipelines using \reader{} and compared the compute requirements of the original pipeline to the optimized version using \reader{}. All experiments in this section are performed on a single machine with 64 GB of DDR5 RAM, a 12-core AMD Ryzen 9 9900X processor, and a 2 TB PCIe 4.0 NVMe SSD.

We focus on reimplementing labeling and featurization for predicting the length of stay on both PyHealth \citep{pyhealth} and MIMIC-IV Data Pipeline \citep{pipeline}. PyHealth and MIMIC-IV Data Pipeline were chosen as they are popular projects that use very different implementation strategies. PyHealth focuses more on Python pickle and CPU efficiency, while MIMIC-IV Data Pipeline relies more heavily on pandas and is more memory efficient. Labeling and featurization were chosen as reimplementation targets as they have concrete input/output expectations that are easy to verify. We perform experiments on the public datasets eICU, MIMIC-III and MIMIC-IV.

One complication with reimplementation experiments is that PyHealth and MIMIC-IV Data Pipeline use different subsets of MIMIC, including different tables and columns within those tables. In order to perform fair comparisons, we convert data from PyHealth and MIMIC-IV Data Pipeline's internal formats to MEDS to make sure that \reader{} processes the same data as the original pipeline. This conversion would not have been necessary if PyHealth and MIMIC-IV Data Pipeline had been designed with MEDS support from the start. This conversion is primarily limited by the efficiency of PyHealth and MIMIC-IV Data Pipeline, but is relatively fast, with performance reported in Appendix \ref{a:convert}. We also need to convert from MEDS to \reader{}'s \texttt{SubjectDatabase}. Converting to a \texttt{SubjectDatabase} is a once-per-dataset cost that is reported in Appendix \ref{a:meds_convert}. This once-per-dataset cost should not be important for repeated analysis on the same dataset, as it will be amortized across many runs, but might be important when trying to run one analysis across many datasets.

We then reimplement labeling and featurization independently for both pipelines, and run the original code and our reimplementation against MIMIC-IV. Unfortunately, we cannot successfully run PyHealth on MIMIC-IV due to out-of-memory issues, so we instead use MIMIC-III and eICU for PyHealth experiments. We use GNU Time \citep{time} to measure the wall clock time and peak memory usage. We measure disk usage by inspecting the cache directories for the pipelines. Appendix \ref{a:benchmark} contains further details and references to the code for these experiments. Table \ref{t:results} contains the time, memory, and disk usage statistics for these experiments.

Our results demonstrate that \reader{} allows us to make these pipelines more efficient on all metrics, sometimes drastically so. The most significant improvements are in the MIMIC-IV Data Pipeline, where a runtime of over 80 hours was reduced to seconds. This speed improvement is primarily due to \reader{}'s O(1) subject lookup as opposed to pandas's $O(n)$ subject lookup, which scales poorly to datasets like MIMIC-IV, which has over 300,000 admissions. We also see considerable improvements in memory for PyHealth, as PyHealth by default stores all subjects in memory while \reader{} only stores one subject in memory at a time.

\section{Discussion And Conclusion}
\label{sec:discussion}

We have presented \reader{}, a Python package that can be used to construct efficient EHR processing pipelines. We have validated the efficiency of our approach by reimplementing sections of major EHR processing pipelines and have shown that our reimplementations are drastically more efficient than the originals, with improvements that are at least one or two orders of magnitude.

We expect there to be three major benefits of \reader{}. First, speeding up EHR processing improves researcher efficiency. The faster research code runs, the faster researchers can iterate on ideas and experiments. Second, \reader{}'s improved efficiency makes it feasible to scale up analysis and machine learning work to massive datasets with hundreds of millions of records. Massive datasets are becoming increasingly important in the era of foundation models where we seek to pretrain large models on oceans of medical data. Finally, \reader{}'s improved efficiency is helpful in deployment scenarios where results need to be returned quickly due to patient care needs.

However, there are some important limitations of \reader{}. First, \reader{} currently supports per-event properties that correspond directly to a single Python type. For example, we support strings, integers, floating point numbers, and datetimes, but do not support lists and structs. Events with list or struct properties require additional handling when using \reader{}, hurting readability and performance. We expect to address this in future versions by adding direct support for those types. The second limitation of \reader{} is that it still relies heavily on Python, as the user-supplied Python transformation functions are run directly in the Python interpreter. The reliance on Python, and the overhead caused by Python, require future work to convert the user-provided transformations from Python to native code to avoid that overhead.  A final limitation of \reader{} is that it depends on the MEDS standard and does not support data elements such as high-time resolution data (such as heart-rate monitors) that are not currently supported by MEDS. However, we expect this issue to be solved as more and more data support is added to MEDS.


\section*{Acknowledgements}

We would also like to thank Hejie Cui for providing thoughtful comments on this paper.
Matthew McDermott gratefully acknowledges support from a Berkowitz Postdoctoral Fellowship. Jason Fries, and Nigam Shah acknowledge support from the Debra and Mark Leslie endowment for AI in healthcare.

\clearpage

\bibliography{jmlr-sample}

\clearpage

\appendix

\onecolumn

\section*{Appendices}

\section{Conversion of PyHealth and MIMIC-IV Data Pipeline to MEDS}\label{a:convert}

One problem with running comparisons against tools like PyHealth and MIMIC-IV Data Pipeline is that every tool processes a different subset of MIMIC, processing different tables and different fields within those tables. In order to make a fair comparison, it is necessary that our \reader{} reimplementation process the same data used in the original pipeline. The most straightforward solution is to convert the intermediates generated by the original pipeline to MEDS so that \reader{} can directly process them. Note that this is just a workaround; if these pipelines were to be fully ported to MEDS and \reader{}, this conversion step would not be necessary. 

The code for our conversion scripts can be found in our code repository, \url{https://github.com/som-shahlab/meds_reader_paper_code/tree/main/conversion/README.md}. We also report the time and resources required for this conversion in Table \ref{t:conversion}, although most of the cost here is due to PyHealth and MIMIC-IV Data Pipeline and not \reader{} or MEDS.

\begin{table*}[!htbp]
\begin{tabular}{cp{5cm}*{4}{c}}
\toprule
Dataset & Source Data Provided By & Memory (GB) & CPU (s) & Disk (GB) \\
\midrule

eICU & PyHealth & 17.20  & 130.27  & 0.062 \\
MIMIC-III & PyHealth & 24.04  & 101.25  & 0.057 \\
MIMIC-IV & MIMIC-IV Data Pipeline & 28.61  & 15.32  & 0.35 \\

\bottomrule

\end{tabular}
\caption{Compute resources required for converting from the source library data format to MEDS. We measure peak memory usage, wall clock CPU time, and disk usage for cache directories.}
\label{t:conversion}
\end{table*}

\section{meds\_reader\_convert cost}\label{a:meds_convert}

In order to use \reader{} on a MEDS dataset, it must be converted to a \reader{} \texttt{SubjectDatabase} using the meds\_reader\_convert program. Conversion to \texttt{SubjectDatabase} is a once-per-dataset cost and is relatively lightweight, but it is important to account for it. We do not include this cost in our main timing tables, as we assume we are in a situation where a dataset is created once, but we include timing information for it here in Table \ref{t:meds_convert}. We do not report disk usage as this process does not use disk outside of the input/output.

\begin{table*}[!htbp]
\begin{tabular}{cp{5cm}*{4}{c}}
\toprule
Dataset & Source Data Provided By & Memory (GB) & CPU (s)  \\
\midrule

eICU & PyHealth & 1.77 GB  & 5.02  \\
MIMIC-III & PyHealth & 0.87  & 4.45  \\
MIMIC-IV & MIMIC-IV Data Pipeline & 1.28 & 19.19 \\

\bottomrule

\end{tabular}
\caption{Compute resources required for converting from MEDS to a \reader{} \texttt{SubjectDatabase}. We measure peak memory usage and wall clock CPU time. }
\label{t:meds_convert}
\end{table*}

\section{Reimplementation Code And Setup}\label{a:benchmark}

We reimplement parts of PyHealth and MIMIC-IV Data Pipeline to verify the performance of \reader{}. To simplify our setup, we explicitly focus on labeling and featurization of a length of stay prediction task. We attempt to configure each pipeline in a manner as simple as possible in accordance with the documentation. The code for these experiments can be found in our code repository at \url{https://github.com/som-shahlab/meds_reader_paper_code/tree/main/reference_implementation/README.md}. For PyHealth, we enable the DIAGNOSES\_ICD, PROCEDURES\_ICD, PRESCRIPTIONS and LABEVENTS tables and define a length of stay task following the example in their tutorial. For the MIMIC-IV Data Pipeline, we select non-ICU data with no disease filtering and use all supported features (diagnosis, labs, procedures, medications). We use their provided length of stay greater than three days task definition and set a prediction time of 24 hours after readmission. 

For both PyHealth and MIMIC-IV Data Pipeline, we reimplement the labeling and featurization for those configuration settings using \reader{}, using the data converted to MEDS as described in Appendix \ref{a:convert}. Our reimplementation code can be found in our code repository at \url{https://github.com/som-shahlab/meds_reader_paper_code/tree/main/reimplementation/README.md}. 

\end{document}